\documentclass[10pt,final,journal]{IEEEtran}

\usepackage{amsmath}
\usepackage{amsfonts}
\usepackage{amssymb}
\usepackage{amsthm}
\usepackage{color}
\usepackage{array}
\usepackage{cite}
\usepackage{enumerate}
\usepackage{graphicx}
\usepackage[hang]{subfigure}
\usepackage{epstopdf}
\usepackage{epsfig}
\usepackage{footmisc}
\usepackage{amsbsy}
\usepackage{bm}
\usepackage{cases}
\usepackage{multirow}
\usepackage{algorithmic}
\usepackage{comment}
\usepackage{times}
\usepackage{paralist}
\usepackage[ruled]{algorithm2e}
\usepackage{setspace}
\usepackage[breakable]{tcolorbox}
\usepackage{kotex}
\usepackage{booktabs}

\newcommand{\bmx}{{\bm x}}

\newcommand{\bmh}{{\bm h}}
\newcommand{\bms}{{\bm s}}

\newcommand{\bma}{{\bm a}}

\newcommand{\bmb}{{\bm b}}

\newcommand{\bmtheta}{\bm{\theta}}

\newcommand{\set}[1]{\ensuremath{\mathcal #1}}
\newcommand{\grad}[1]{\nabla #1}

\begin{document}
\bstctlcite{IEEEexample:BSTcontrol}

\title{Spiking Neural Networks -- \\ Part II: Detecting Spatio-Temporal Patterns}

%
%


\author{Nicolas~Skatchkovsky$^\ast$, Hyeryung~Jang$^\ast$ and Osvaldo~Simeone
\thanks{$^{\ast}$ The first two authors have equally contributed to this work. The authors are with the Centre for Telecommunications Research, Department of Engineering, King’s College London, United Kingdom. (e-mail: \{nicolas.skatchkovsky, hyeryung.jang, osvaldo.simeone\}@kcl.ac.uk). 
This work has received funding from the European Research Council (ERC) under the European Union's Horizon 2020 Research and Innovation Programme (Grant Agreement No. 725731).}
}


\maketitle

\begin{abstract}
Inspired by the operation of biological brains, Spiking Neural Networks (SNNs) have the unique ability to detect information encoded in spatio-temporal patterns of spiking signals. Examples of data types requiring spatio-temporal processing include logs of time stamps, e.g., of tweets, and outputs of neural prostheses and neuromorphic sensors. In this paper, the second of a series of three review papers on SNNs, we first review models and training algorithms for the dominant approach that considers SNNs as a Recurrent Neural Network (RNN) and adapt learning rules based on backpropagation through time to the requirements of SNNs. In order to tackle the non-differentiability of the spiking mechanism, state-of-the-art solutions use surrogate gradients that approximate the threshold activation function with a differentiable function. Then, we describe an alternative approach that relies on probabilistic models for spiking neurons, allowing the derivation of local learning rules via stochastic estimates of the gradient. Finally, experiments are provided for neuromorphic data sets, yielding insights on accuracy and convergence under different SNN models.
\end{abstract}

\vspace{-3pt}

\section{Introduction}
Spiking Neural Networks (SNNs) have recently emerged as a more energy efficient alternative to Artificial Neural Networks (ANNs) for applications involving on-device learning and inference. SNNs are inspired by the operation of biological brains, and function using sparse, binary \textit{spiking} signals. Unlike the static, real-valued, signals used to train ANNs, spiking signals encode information using the \textit{spatio-temporal} patterns defined by spike timings. Thanks to efficient hardware implementations, SNNs often may unlock a new class of applications for mobile or embedded devices. 

Training ANNs relies on the backpropagation (BP) algorithm, which carries out ``credit assignment'' for the output loss to the network by implementing the chain rule of differentiation. Unlike for ANNs, the gradient of the output loss does not provide an informative learning signal for SNNs. In fact, spiking neurons are characterized by threshold activation function, whose derivative is zero almost everywhere. Different lines of research have been pursued to overcome this challenge. As we reviewed in Part I, a first approach bypasses the problem by training an ANN for a specific task, and then converting its weights to a SNN at inference time. When employing such techniques, SNNs are used to detect spatial patterns from data, hence implementing the functionality of ANNs \cite{rueckauer2017convrate, cao2015convcnn, comsa2020temporal}, albeit with energy consumption improvements.

A major argument in favor of the use of SNNs is their ability to process information encoded both over space and over time 
(see Fig.~\ref{fig:snn_encoding}). The slow progress in the development of training algorithms for SNNs that are able to detect spatio-temporal patterns stems in part of the lack of widely accepted benchmark \textit{neuromorphic} data sets \cite{davies2019spikemark}. The recent emergence of data sets captured using neuromorphic sensors \cite{davies2019spikemark, Lichtsteiner2006dvs_camera, amir2017dvsgesture, cramer2020heidelberg} is quickly changing the research landscape, leading to a number of breakthroughs, while still leaving important open questions. 

Training SNNs for detecting spatio-temporal patterns has followed into two main approaches. The first, currently dominant, class of methods views SNNs as a subclass of Recurrent Neural Networks (RNNs), and adapts standard algorithms for RNNs such as backpropagation through time (BPTT) and forward-mode differentiation to the unique features of SNNs. In order to tackle the issues due to non differentiable activation function, state-of-the-art solutions are given by surrogate gradient (SG) algorithms \cite{neftci2019surrogate} that smooth out the derivative with a differentiable surrogate function, and/or by approximating the learning signals with either random feedback signals \cite{neftci2019surrogate, zenke2018superspike} or with locally computed surrogate losses \cite{kaiser2020decolle}.

The second line of work characterizes the operation of SNNs in a probabilistic manner by introducing randomness in the spiking mechanism \cite{pillow08:spatio, gerstner2002spiking, jang19:spm}. A probabilistic model defines differentiable output loss and enables the direct derivation of learning algorithms using principled learning criteria, which do not require the implementation of backpropagation mechanisms. 


The remainder of this paper is organized as follows. In Sec.~\ref{sec:spatio-temporal}, we provide a review of SNN models used for detecting spatio-temporal patterns, then introduce the corresponding learning algorithms in Sec.~\ref{sec:spatio-temporal-alg}. In Sec.~\ref{sec:spatio-temporal-exp}, we explore an application for classification task. In conclusion, we introduce extensions in Sec.~\ref{sec:conclusion}.

\vspace{-3pt}
\section{Detecting Spatio-Temporal Patterns: Models} \label{sec:spatio-temporal}

\begin{figure}[t!]
\centering
\includegraphics[height=0.27\columnwidth]{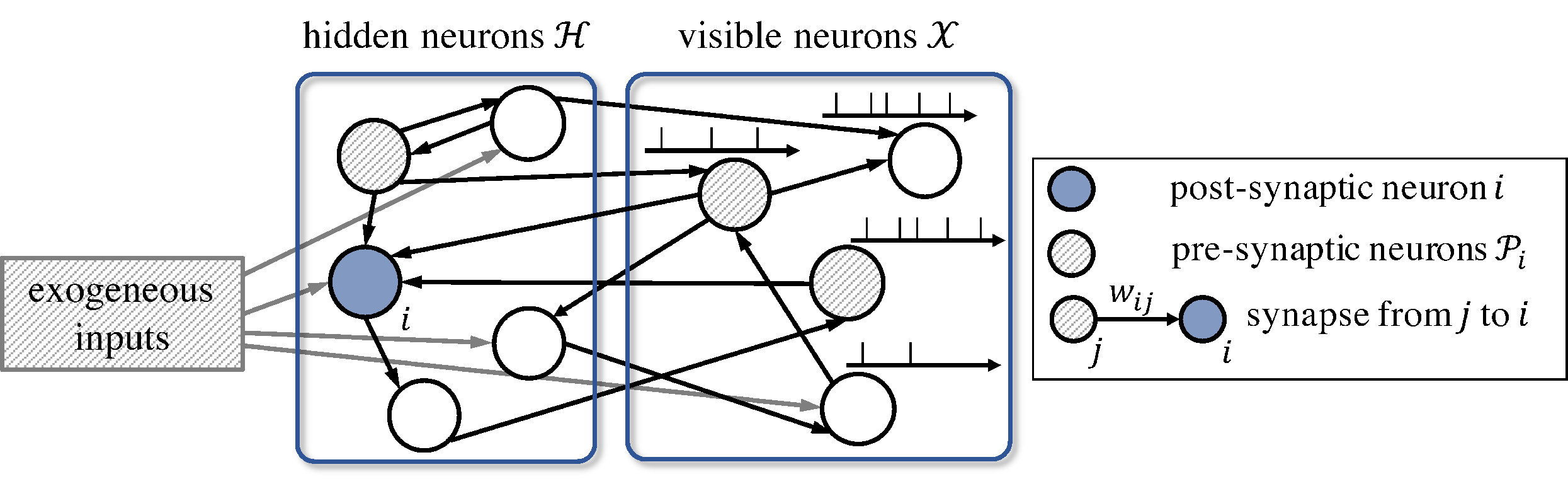}
\vspace{-0.2cm}
\caption{Architecture of an SNN with exogeneous inputs, $|\set{X}| = 4$ visible neurons, and $|\set{H}| = 5$ hidden neurons. Synapses are represented by directed links between two neurons.
} 
\label{fig:snn_glm}
\vspace{-0.4cm}
\end{figure}

In this section, we first review the details of the deterministic discrete-time Spike Response Model (SRM) introduced in Part I, and then describe a probabilistic extension of the SRM known as Generalized Linear Model (GLM). Throughout this paper, we consider a general SNN architecture described by a directed, possibly cyclic, graph with spiking neurons as vertices and synapses as directed edges. As seen in Fig.~\ref{fig:snn_glm}, each neuron $i$ receives synaptic connections from a subset $\set{P}_i$ of other spiking neurons. With a slight abuse of notations, we take the set $\set{P}_i$ to include also exogeneous inputs. 

\vspace{-0.25cm}
\subsection{Deterministic Spike Response Model (SRM)}

As described in Part I, under the discrete-time SRM, the binary output $s_{i,t} \in \{0,1\}$ of spiking neuron $i \in \set{V}$ at time $t$ depends on its membrane potential $u_{i,t}$ as
\begin{align} \label{eq:srm-threshold}
    s_{i,t} = \Theta(u_{i,t} - \vartheta),
\end{align}
where $\Theta(\cdot)$ is the Heaviside step function. Accordingly, a spike is emitted when the membrane potential $u_{i,t}$ crosses a fixed threshold $\vartheta$, where the membrane potential is obtained as 
\begin{align} \label{eq:potential-recur}
    u_{i,t} = \sum_{j \in \set{P}_i} w_{ij} \big( \underbrace{\alpha_t \ast s_{j,t}}_{:=~ p_{j,t}} \big) + \big( \underbrace{\beta_t \ast s_{i,t}}_{:=~ r_{i,t}} \big) + \gamma_i,
\end{align}
where $w_{ij}$ is the synaptic weight from pre-synaptic neuron $j \in \set{P}_i$ to post-synaptic neuron $i$; $\alpha_t$ and $\beta_t$ represent the synaptic and feedback spike responses, respectively; and $\ast$ denotes the convolution operator. We collect in vector $\bmtheta = \{\bmtheta_i\}_{i \in \set{V}}$ the model parameters, with $\bmtheta_i := \{w_{ij}\}_{j \in \set{P}_i}$ being the local model parameters of each neuron $i$. The bias is typically set to $\gamma_i = 0$.

SRM is commonly implemented with the \textit{alpha-function} spike response $\alpha_t = \exp(-t/\tau_{\text{mem}}) - \exp(-t/\tau_{\text{syn}})$ and the exponentially decaying feedback filter $\beta_t = -\exp(-t/\tau_{\text{ref}})$ for $t \geq 1$ with some positive constants $\tau_{\text{mem}}, \tau_{\text{syn}}$, and $\tau_{\text{ref}}$. 
With these choices, the filtered contributions in \eqref{eq:potential-recur} can be computed using the recursive equations 
\begin{subequations} \label{eq:filter-recur}
\begin{align} 
    p_{j,t} &= \exp\big( -1/\tau_{\text{mem}}\big) p_{j,t-1} + q_{j,t-1}, \label{eq:filter-recur-a} \\ 
    \text{with}~~ q_{j,t} &= \exp\big( -1/\tau_{\text{syn}}\big) q_{j,t-1} + s_{j,t-1}, \label{eq:filter-recur-b} \\
    \text{and}~~ r_{i,t} &= \exp\big( -1/\tau_{\text{ref}}\big) r_{i,t-1} + s_{i,t-1}. \label{eq:filter-recur-c}
\end{align}
\end{subequations}
Using \eqref{eq:filter-recur-a}-\eqref{eq:filter-recur-b}, each synapse $(j,i)$ between pre-synaptic neuron $j$ and post-synaptic neuron $i$ computes the synaptic filtered trace $p_{j,t}$ in \eqref{eq:potential-recur} via a second-order autoregressive (AR) filter, while the soma of each neuron $i$ computes feedback filtered trace $r_{i,t}$ in \eqref{eq:potential-recur} via a first-order AR filter. The dynamics of spiking neurons \eqref{eq:srm-threshold}-\eqref{eq:filter-recur} define the operation of a Recurrent Neural Network (RNN) with threshold activation function \eqref{eq:srm-threshold} and binary activations \cite{neftci2019surrogate, wozniak2020snu}.

\vspace{-0.25cm}
\subsection{Probabilistic Generalized Linear Model (GLM)}
\label{sec:glm}

A probabilistic generalization of the SRM used in computational neuroscience is the Generalized Linear Model (GLM), which introduces randomness in the spiking mechanism \eqref{eq:srm-threshold}. Define as $\bms_t = (s_{i,t}: i \in \set{V})$ the spiking signals for all neurons at time $t$ and as $\bms_{\leq t-1} = (\bms_1, \ldots, \bms_{t-1})$ all signals up to time $t-1$. Under GLMs, the spiking probability of neuron $i$ conditioned on $\bms_{\leq t-1}$ is given as \cite{jang19:spm, gerstner2002spiking, pillow08:spatio}
\begin{align} \label{eq:glm-prob}
    p_{\bmtheta_i}(s_{i,t} = 1 | \bms_{\leq t-1}) = p_{\bmtheta_i}(s_{i,t} = 1 | u_{i,t}) = \sigma_\Delta(u_{i,t}),
\end{align}
where $\sigma_\Delta(x)=1/(1+\exp(- \Delta x))$ is the sigmoid function, $\Delta > 0$ is a bandwidth parameter, and the membrane potential $u_{i,t}$ is defined in \eqref{eq:potential-recur}. Conditioned on $\bms_{\leq t-1}$, the variables $s_{i,t}$ are independent across the neuron index $i$. As per \eqref{eq:glm-prob}, a larger membrane potential increases the spiking probability. The notation used in \eqref{eq:glm-prob} emphasizes the dependence on the local model parameters $\bmtheta_i = \{\{w_{ij}\}_{j \in \set{P}_i}, \gamma_i\}$ of neuron $i$. The GLM \eqref{eq:glm-prob} reduces to the SRM model \eqref{eq:srm-threshold} when $\Delta \rightarrow \infty$ and $\gamma_i = -\vartheta$ so that the sigmoid function $\sigma_\Delta(u_{i,t})$ tends to the hard threshold function $\Theta(u_{i,t} - \vartheta)$. Throughout the paper, we simply denote the sigmoid function by $\sigma(\cdot)$.

By \eqref{eq:glm-prob}, the negative log-probability of the spike is given by 
\begin{align}
    - \log p_{\bmtheta_i}(s_{i,t} | u_{i,t}) = \ell\big( s_{i,t}, \sigma(u_{i,t})\big),
\end{align}
where $\ell(a,b) = -a \log(b) - (1-a)\log(1-b)$ is the binary cross-entropy loss function. Using the chain rule, the negative log-probability of the spike sequence $\bms_{\leq T}$ emitted by all neurons up to time $T$ can be written as $- \log p_{\bmtheta}(\bms_{\leq T}) = \sum_{t=1}^T \sum_{i \in \set{V}} \ell\big( s_{i,t}, \sigma(u_{i,t})\big)$. 
As a remark, in GLMs, it is common to associate each synapse with multiple spike responses and corresponding weights, allowing distinct temporal receptive fields to be learned across the spiking neurons \cite{pillow08:spatio, jang19:spm}. We also note that an extension of the GLM model with non-binary neurons can be found in \cite{jang20:vowel}.

\begin{figure*}[t!]
\centering
\subfigure[]{
\includegraphics[height=0.365\columnwidth]{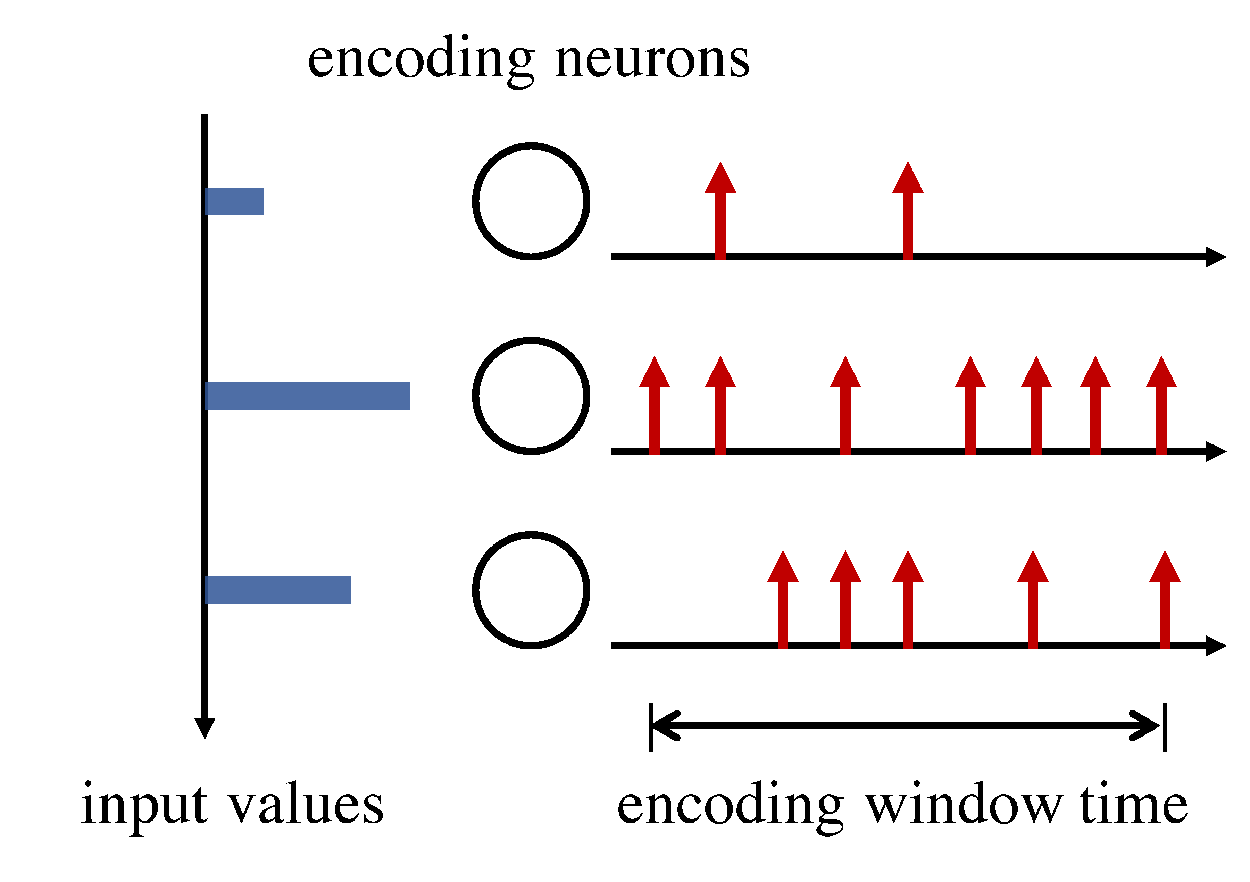} \label{fig:rate_encoding}
} 
\hspace{-0.9cm}
\subfigure[]{
\includegraphics[height=0.365\columnwidth]{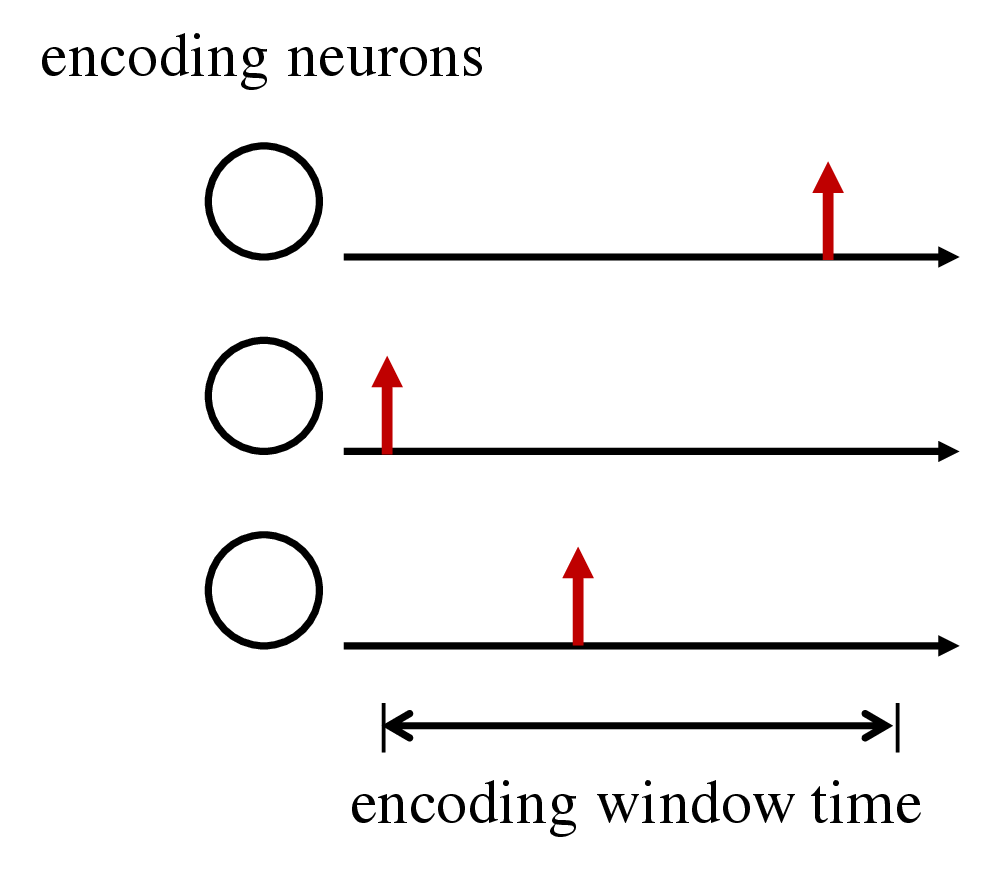} \label{fig:temporal_encoding}
}
\hspace{-0.55cm}
\subfigure[]{
\includegraphics[height=0.435\columnwidth]{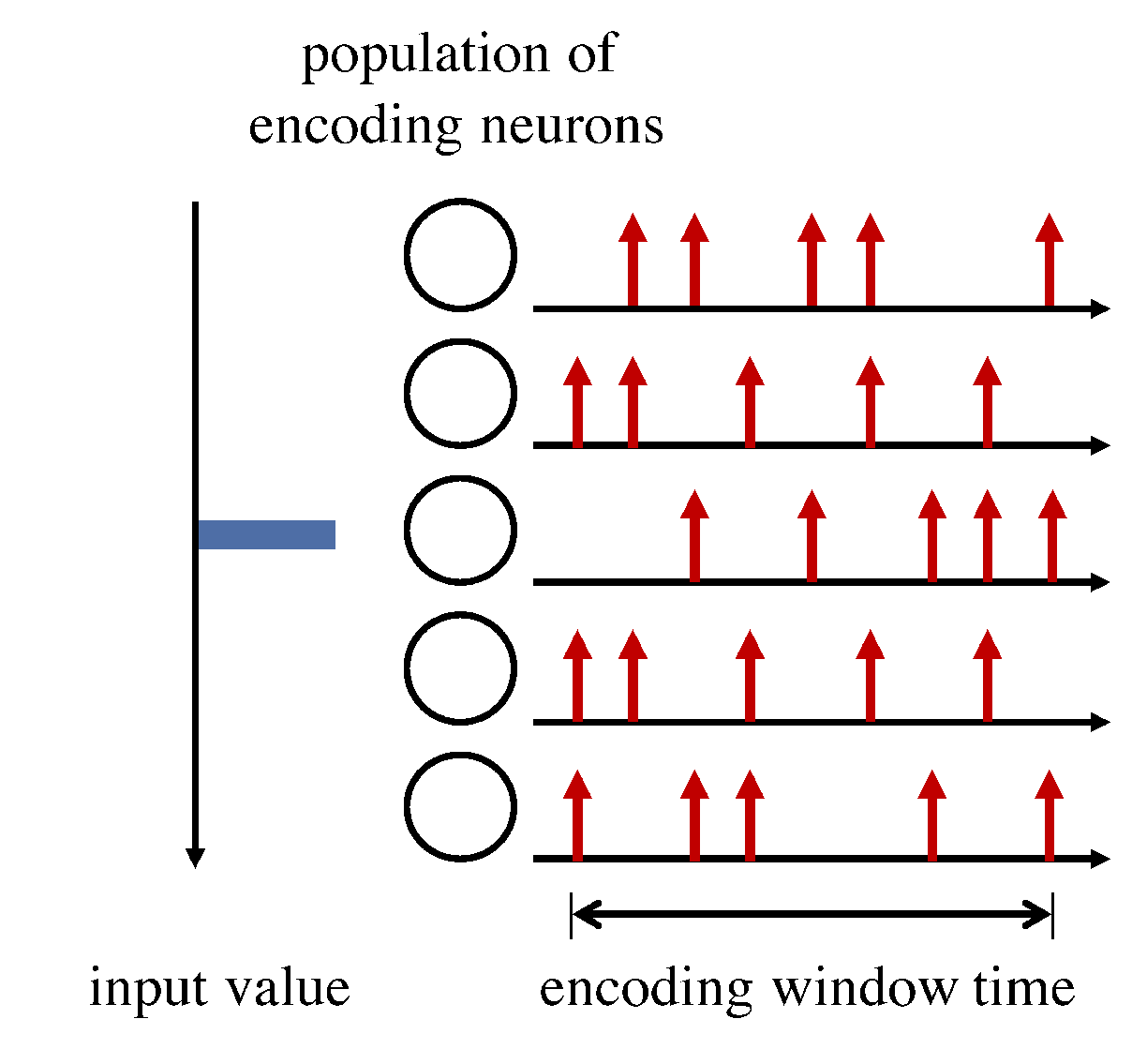} \label{fig:rate_encoding_population}
}
\hspace{-0.45cm}
\subfigure[]{
\includegraphics[height=0.435\columnwidth]{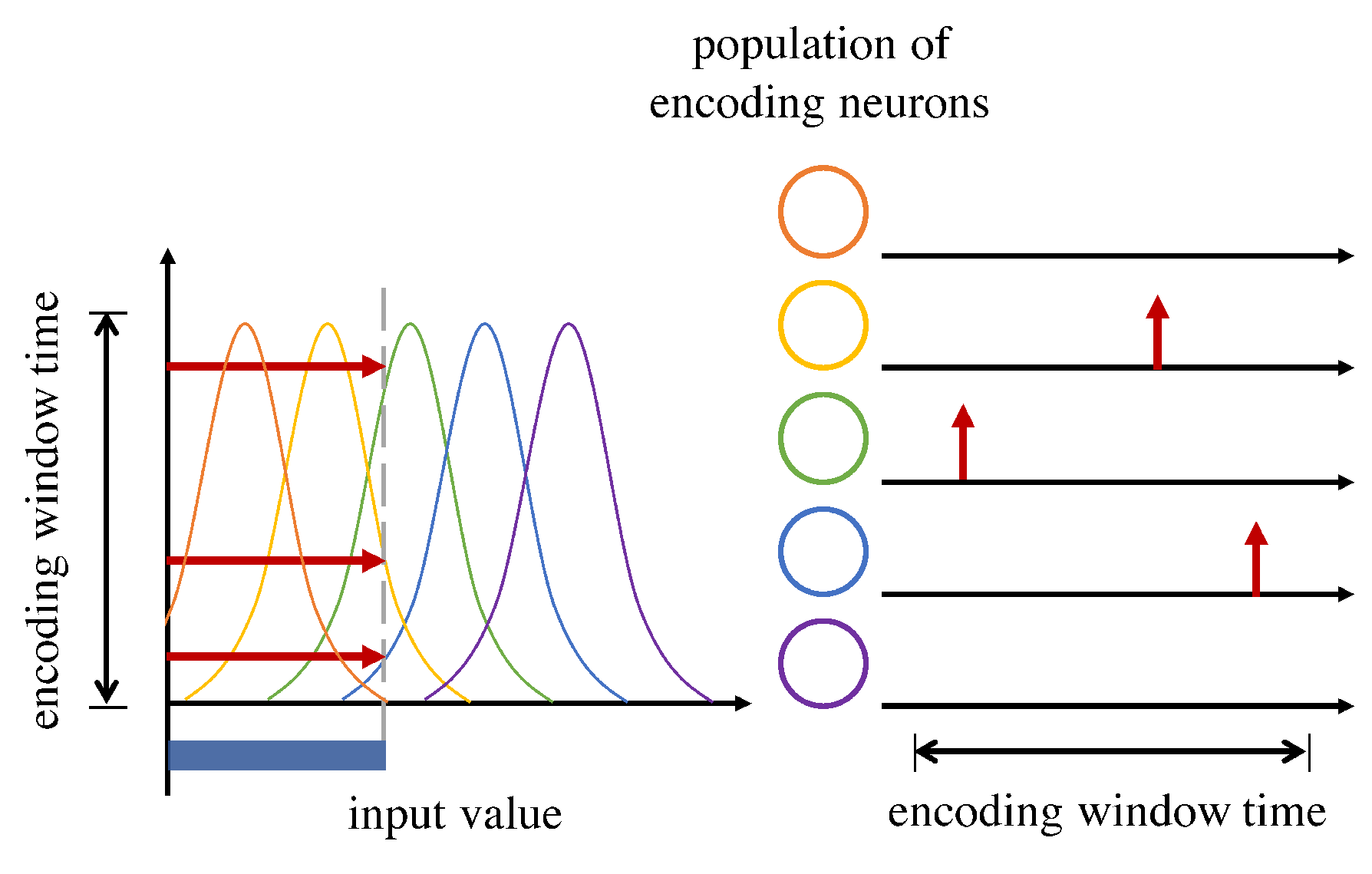} \label{fig:temporal_encoding_population}
}
\caption{Illustration of rate, time, and population encoding principles: (a) With rate coding, each input value, as shown using the horizontal blue bars on the left, is encoded in the spike rate of the corresponding encoding neuron: A larger input generates a large number of spikes within a fixed encoding window time; (b) With time encoding, an input value is encoded by the timing of one or more spikes emitted by the corresponding encoding neuron; (c) With population rate encoding, an input value is encoded by the average spike rate for a set of encoding neurons; (d) With population time encoding, an input value is encoded by at most one spike for each of the encoding neurons, whose timings are given by a ``receptive field'' function shown on the left \cite{bohte2002unsupervised}.
} 
\label{fig:snn_encoding}
\vspace{-0.4cm}
\end{figure*}


For any two jointly distributed random processes $\bma_{\leq T}, \bmb_{\leq T}$, we introduce causal conditioning notation \cite{kramer1998directed} of $\bma_{\leq T}$ given $\bmb_{\leq T}$ denoted by $p(\mathbf{a}_{\leq T} ||\mathbf{b}_{\leq T}) = \prod_{t=1}^{T} p(\mathbf{a}_{t} | \mathbf{a}_{\leq t - 1}, \mathbf{b}_{\leq t})$.

\section{Detecting Spatio-Temporal Patterns: Algorithms}
\label{sec:spatio-temporal-alg}

In this section, we consider the problem of training SNNs using either supervised or unsupervised learning for the purpose of detecting, or generating, spatio-temporal patterns of spiking signals. Theses are encoded in target spiking outputs for a subset $\set{X} \subseteq \set{V}$ of neurons, referred to as ``visible'' (see Fig.~\ref{fig:snn_glm}). 
In supervised learning, visible neurons represent the read-out layer that encodes the index of the desired class for classification or a continuous target variable for regression. Information encoding into spiking signals may be based on rate, time, and population encoding principles (see Fig.~\ref{fig:snn_encoding}). In unsupervised generative learning, the output of the visible neurons directly provides the desired signals to be generated by the SNN.

Training relies on the availability of a training data set $\set{D}$ containing $|\set{D}|$ examples of spiking signals $\bmx_{\leq T}$ of duration $T$ for visible neurons $\set{X}$. The spiking signals $\bmh_{\leq T}$ of the complementary subset $\set{H} = \set{V} \setminus \set{X}$ of \textit{hidden}, or latent, neurons are not specified by the training set, and their behavior should be adapted during training to ensure the desired behavior of visible neurons. The error between the target spiking signals $\bmx_{\leq T} \in \set{D}$ and the actual, possibly probabilistic, output spiking signals produced by the SNN is measured by a loss function that is defined differently for SRMs and GLMs. Defining as $\set{L}_{\bmx_{\leq T}}(\bmtheta)$ the loss measured on the example $\bmx_{\leq T} \in \set{D}$ for an SNN with model parameters $\bmtheta$, standard stochastic gradient-based learning rules update the model parameters $\bmtheta$ in the opposite direction of the gradient $\grad_{\bmtheta} \set{L}_{\bmx_{\leq T}}(\bmtheta)$ as $\bmtheta \leftarrow \bmtheta - \eta \grad_{\bmtheta} \set{L}_{\bmx_{\leq T}}(\bmtheta)$, where $\eta$ is the learning rate.

\vspace{-0.2cm}
\subsection{Training SRM-Based Models}
\label{sec:srm}
As compared to conventional methods for ANNs, applying gradient-based rules to training SRM-based SNNs poses a number of new challenges. First, the derivative of the activation function \eqref{eq:srm-threshold} is zero almost everywhere, making the gradient not informative for training. Second, as seen, SNNs are recurrent architectures due to refractoriness mechanism and, possibly, to cyclic directed topologies. SNNs hence inherit from Recurrent Neural Networks (RNNs) the problem of carrying out spatial (across neurons) and temporal (across time) {\em credit assignment} for the training loss. Finally, as required in most neuromorphic hardware platforms, training algorithms should only involve {\em local operations} along with minimal global signaling, making the standard backpropagation through time (BPTT) RNN training algorithm not desirable for SNNs.

A state-of-the-art approach to tackle the first problem is given by surrogate gradient (SG) methods. SG algorithms smooth out the Heaviside function in \eqref{eq:srm-threshold} when computing the gradient $\grad_{\bmtheta} \set{L}_{\bmx_{\leq T}}(\bmtheta)$. To elaborate, assume that the loss function is written as a sum over $t=1,\ldots,T$ and $i \in \set{X}$ as
\begin{align} \label{eq:loss-decomp}
    \set{L}_{\bmx_{\leq T}}(\bmtheta) = \sum_{t=1}^T \sum_{i \in \set{X}} L(x_{i,t}, \bar{x}_{i,t}),
\end{align}
where each term $L(x_{i,t},\bar{x}_{i,t})$ is a local loss measure for visible neuron $i \in \set{X}$ at time $t$ that depends only on the target output $x_{i,t}$ of neuron $i$ at time $t$ and on the actual outputs $\bar{x}_{i,t}$ of the same neuron. Note that the output $\bar{x}_{i,t}$ depends on the model parameters $\bmtheta$ (from \eqref{eq:srm-threshold}-\eqref{eq:filter-recur}).
With this decomposition \eqref{eq:loss-decomp}, the partial derivative of the training loss $\set{L}_{\bmx_{\leq T}}(\bmtheta)$ with respect to a synaptic weight $w_{ij}$ can be approximated as
\begin{align} \label{eq:loss-der}
    \frac{\partial \set{L}_{\bmx_{\leq T}}(\bmtheta)}{\partial w_{ij}} \approx \sum_{t=1}^T \underbrace{ \sum_{k \in \set{X}} \frac{\partial L(x_{k,t},\bar{x}_{k,t})}{\partial s_{i,t}}}_{e_{i,t}} \cdot \underbrace{\frac{\partial s_{i,t}}{\partial u_{i,t}}}_{\Theta'(u_{i,t}-\vartheta)} \cdot \underbrace{\frac{\partial u_{i,t}}{\partial w_{ij}}}_{\alpha_t \ast s_{j,t}},
\end{align}
where the first term $e_{i,t}$ is the derivative of the loss at time $t$ with respect to the post-synaptic neuron $i$'s output at time $t$; the second term is problematic since it includes the derivative of the threshold function $\Theta'(\cdot)$ from \eqref{eq:srm-threshold}, which is zero almost everywhere; and the third term can be directly computed from  \eqref{eq:filter-recur} as the filtered pre-synaptic trace $\partial u_{i,t}/ \partial w_{ij} = p_{j,t} = \alpha_t \ast s_{j,t}$. The approximation stems from ignoring the contribution of the derivatives \big($\sum_{k \in \set{X}} \frac{\partial L(x_{k,t}, \bar{x}_{k,t})}{\partial s_{i,t'}}$\big) with $t' < t$. Note that in fact, the synaptic weight $w_{ij}$ affects also the output $s_{i,t'}$ for $t'<t$. Following the SG approach, the derivative $\Theta'(\cdot)$ is substituted with the derivative of a differentiable surrogate function, such as rectifier \cite{bohte2002spikeprop}, sigmoid \cite{zenke2018superspike, bellec2018long}, or exponential function \cite{shrestha2018slayer}. For example, with a sigmoid surrogate, we have $\partial \bar{x}_{i,t}/\partial u_{i,t} \approx \sigma'(u_{i,t}-\vartheta)$.

From \eqref{eq:loss-der}, the resulting SG-based rule is given as
\begin{align} \label{eq:sg-update}
    w_{ij} \leftarrow w_{ij} - \eta \sum_{t=1}^T \Delta w_{ij,t}.
\end{align}
with
\begin{align} \label{eq:srm-update}
    \Delta w_{ij,t} = \underbrace{e_{i,t}}_{\text{error signal}} \underbrace{\sigma'(u_{i,t}-\vartheta)}_{\text{post}_{i, t}} \big( \underbrace{\alpha_t \ast s_{j,t}}_{\text{pre}_{j, t}} \big).
\end{align}
The update \eqref{eq:sg-update} implements a {\em three-factor} rule \cite{fremaux2016neuromodulated} in that the contribution $\Delta w_{ij,t}$ of each time $t$ to the update of a synaptic weight $w_{ij}$ is given by the product of a local pre-synaptic term $\text{pre}_{j,t}$, which is a function of the activity of the pre-synaptic neuron $j$; a local post-synaptic term $\text{post}_{i,t}$, which measures the current ``sensitivity'' to changes in the membrane potential of post-synaptic neuron $i$; and a \textit{per-neuron} error signal $e_{i, t}$ that depends on the overall operation of the system. The computation of this term generally requires BPTT, or, alternatively, forward-mode differentiation \cite{neftci2019surrogate}. Effective state-of-the-art solutions tackle the second and third challenges mentioned earlier by approximating the error $e_{i,t}$ with either random projections of the error signals measured at the visible neurons \cite{neftci2019surrogate, zenke2018superspike} or with locally computed error signals \cite{kaiser2020decolle}.  

\vspace{-0.15cm}
\subsection{Training GLM-Based SNNs}

The most common learning criterion for probabilistic models is Maximum Likelihood (ML) \cite{koller2009probabilistic}. ML selects model parameters $\bmtheta$ that maximize the log-probability that the set $\set{X}$ of visible neurons outputs desired spiking signals $\bmx_{\leq T} \in \set{D}$. Equivalently, the training loss $\set{L}_{\bmx_{\leq T}}(\bmtheta)$ is given by the negative log-likelihood (NLL) $\set{L}_{\bmx_{\leq T}}(\bmtheta) = - \log p_{\bmtheta}(\bmx_{\leq T}) = - \log \sum_{\bmh_{\leq T}} p_{\bmtheta}(\bmx_{\leq T}, \bmh_{\leq T})$ that requires to average out the stochastic signals $\bmh_{\leq T}$ produced by the hidden neurons. Using the causal conditioning notation (see Sec. \ref{sec:glm}), 
the loss is given as 
\begin{align} \label{eq:marginal-nll}
    \set{L}_{\bmx_{\leq T}}(\bmtheta) = -\log \mathbb{E}_{p_{\bmtheta}(\bmh_{\leq T}||\bmx_{\leq T})} \big[ p_{\bmtheta}(\bmx_{\leq T}||\bmh_{\leq T-1}) \big].
\end{align}
The expectation in \eqref{eq:marginal-nll} causes two problems. First, this expectation is generally intractable since its complexity is exponential in the product $|\mathcal{H}|T$. Second, the loss $\set{L}_{\bmx_{\leq T}}(\bmtheta)$ cannot be factorized over neurons $i \in \mathcal{X}$ and time $t$ as in \eqref{eq:loss-decomp}. 

To cope with outlined problems, a common approach is to upper bound the NLL using Jensen's inequality as
\begin{eqnarray} \label{eq:elbo}
    &&\set{L}_{\bmx_{\leq T}}(\bmtheta) \leq \mathbb{E}_{p_{\bmtheta}(\bmh_{\leq T}||\bmx_{\leq T})}\Bigg[ \sum_{t=1}^T \sum_{i \in \set{X}} \ell\big( x_{i,t}, \sigma(u_{i,t})\big) \Bigg] \cr
    &&\qquad \quad := L_{\bmx_{\leq T}}(\bmtheta),
\end{eqnarray}
where the contribution to the loss for each visible neuron $i \in \set{X}$ at time $t$ is measured locally by the cross entropy $\ell\big(x_{i,t}, \sigma(u_{i,t})\big)$. Note that the membrane potential $u_{i,t}$ generally depends on the output of the hidden neurons. An empirical estimate of the gradient $\grad_{\bmtheta} L_{\bmx_{\leq T}}(\bmtheta)$ can be now obtained by replacing the expectation in \eqref{eq:elbo} with a single sample of the hidden neurons' signals from the distribution $p_{\bmtheta}(\bmh_{\leq T}||\bmx_{\leq T})$. 

For each visible neuron $i \in \set{X}$, the resulting learning rule for synaptic weight $w_{ij}$ can be directly computed as \eqref{eq:sg-update}, with
\begin{subequations} \label{eq:glm-update}
\begin{align} \label{eq:glm-update-vis}
    \Delta w_{ij, t} = \big( \underbrace{ x_{i,t} - \sigma(u_{i,t})}_{\text{post}_{i, t}} \big) \big( \underbrace{\alpha_t \ast s_{j,t}}_{\text{pre}_{j, t}} \big).
\end{align}
For each hidden neuron $i \in \set{H}$, one needs to deal with the additional complication that the distribution of the hidden neurons' outputs $p_{\bmtheta}(\bmh_{\leq T} || \bmx_{\leq T})$ depend on the parameter of the hidden neurons, preventing a direct empirical approximation as in \eqref{eq:glm-update-vis}. To address this issue, we can resort to one of several gradient estimates \cite{mnih14:nvil, aueb2015local}. A notable example is the REINFORCE gradient, which yields
\begin{align} \label{eq:glm-update-hid}
    \Delta w_{ij, t} = \underbrace{\bar{e}_t}_{\text{global error}} \big( \underbrace{h_{i,t}-\sigma(u_{i,t})}_{\text{post}_{i,t}} \big) \big( \underbrace{\alpha_t \ast s_{j,t}}_{\text{pre}_{j,t}} \big),
\end{align}
\end{subequations}
where the update \eqref{eq:glm-update-vis} for visible neurons yields a two-factor rule in that the contribution $\Delta w_{ij,t}$ of each time $t$ is given by the product of a pre-synaptic trace $p_{j,t} = \alpha_t \ast s_{j,t}$, and a post-synaptic term $x_{i,t} - \sigma (u_{i,t})$. Unlike the sensitivity term in \eqref{eq:srm-update}, the post-synaptic term in \eqref{eq:glm-update-vis} provides an error measure between the desired target and the average model output. In contrast, update $\Delta w_{ij,t}$ in \eqref{eq:glm-update-hid} yields a three-factor rule, as it involves also a \textit{common} error signal $\bar{e}_t:= \sum_{i \in \set{X}} \ell\big( x_{i,t}, \sigma(u_{i,t})\big)$. The post-synaptic term $\text{post}_{i,t}$ in \eqref{eq:glm-update-hid} for the hidden neurons can be interpreted as a predictive error between the best (in the least square sense) prediction of the model, namely $\sigma(u_{i,t})$, and the actual realized output $h_{i,t}$ of neuron $i \in \mathcal{H}$ at time $t$. Furthermore, the update \eqref{eq:glm-update-hid} for hidden neurons contains the error signal $\bar{e}_t$ that is necessary to guide the update of hidden neurons given the lack of a specific target for their outputs in the data.

Comparing the updates \eqref{eq:srm-update} and \eqref{eq:glm-update} for SRM and GLM, respectively, we highlight several differences. First, the operation of the system is probabilistic for GLM and deterministic for SRM. Second, as mentioned, the post-synaptic terms in \eqref{eq:glm-update} represent error signals rather than sensitivity measures. Third, only the hidden neurons implement three-factor rules. Fourth, the error signal in \eqref{eq:srm-update} is individual, i.e., per neuron, while it is common to all hidden neurons in \eqref{eq:glm-update-hid}. The presence of a per-neuron error $e_{i,t}$ in \eqref{eq:srm-update}, instead of a global error $\bar{e}_t$ in \eqref{eq:glm-update}, may result in a faster convergence for \eqref{eq:srm-update} if the error signal $e_{i,t}$ is more informative about the derivative of the loss \cite{neftci2019surrogate}. However, an exact evaluation of the error $e_{i,t}$ requires complex mechanisms, such as BPTT, and the effectiveness of the discussed approximations should be evaluated on a case-by-case basis.

\vspace{-0.2cm}

\section{Examples}
\label{sec:spatio-temporal-exp}

\begin{table}[t!]
\caption{Test accuracy of probabilistic SNNs and DECOLLE on MNIST-DVS. Period refers to the inverse of the sampling rate. The asterisk $\ast$ indicates a convolutional architecture.}
\vspace{-0.3cm}
\label{tab:comparison-mnist}
\begin{center}
\begin{small}
\begin{sc}
\begin{tabular}{lcccr}
\toprule
Model & Period & $N_H$ & Input & Acc. \\
\midrule
 & 1 ms & 39,232* & Per-sign & $99.4$\%  \\
 & 1 ms & 19,616* & Binary & $98.9$\%  \\
DECOLLE & 1 ms & $512$ & Per-sign & $86.8$\%  \\
 & 10 ms & $256$ & Per-sign & $73.8$\% \\
 & 25 ms & $256$ & Per-sign & $65.8$\% \\
\midrule
& 25 ms & 512 & Per-sign & $83.50\%$ \\
SNN & 25 ms & 512 & Binary &$80.80\%$ \\
& 25 ms & 256 & Per-sign & $82.80\%$ \\
& 25 ms & 256 & Binary &$79.3\%$ \\
\bottomrule
\end{tabular}
\end{sc}
\end{small}
\end{center}
\vspace{-0.6cm}
\end{table}

In this section, we  compare SRM and GLM-based models in terms of their capacity to classify spatio-temporal patterns. To this end, we focus on the neuromorphic datasets MNIST-DVS \cite{serrano2015poker} and DVSGesture \cite{amir2017dvsgesture}. The former consists of the recordings via a neuromorphic camera of $10,000$ images from the MNIST dataset displayed on a moving screen. Changes in the intensity of the scene result in pixels taking values in the set $\{-1, \ 0, \ +1\}$, where ``$-1$'' indicates a decrease in luminosity, ``$+1$'' an increase, and ``$0$'' no change. We crop the images to $26 \times 26$ pixels in the center of the images. This produces $676$ input spiking signals when we discard information about the polarity of the spikes; or $1,352$ signals when we include two separate inputs per pixel for positive and negative spikes. The duration of the examples is limited to $2$ seconds, and various subsampling periods are considered. The DVSGesture dataset consists of 11 hand gestures displayed to a neuromorphic camera in different lighting conditions. In this case, images are reduced to $32 \times 32$ pixels, yielding $1,024$ input spiking signals when discarding polarity, or $2,048$ with per-sign inputs.

\begin{figure}[t!]
\centering
\includegraphics[width=0.9\columnwidth]{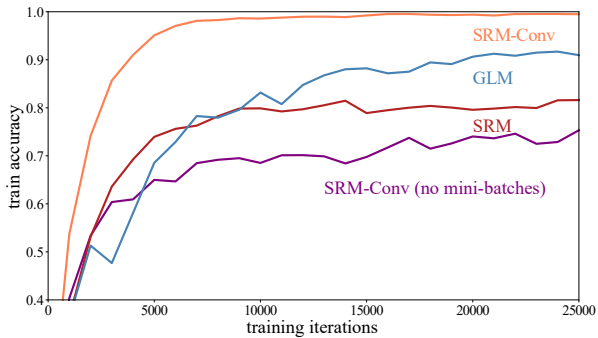} \label{fig:train_acc}
\vspace{-0.2cm}
\caption{Convergence speed of probabilistic GLM and SRM models, with the latter trained via trained with DECOLLE. For SRM-based SNNs, we include results with the convolutional architecture used in \cite{kaiser2020decolle}.
} 
\label{fig:train_convergence}
\vspace{-0.4cm}
\end{figure}

To train SRM-based models, we adopt the surrogate gradient scheme DECOLLE \cite{kaiser2020decolle}, which uses per-layer random error signals as discussed in Sec. \ref{sec:srm}; while GLM-based SNNs are trained as explained in Sec. \ref{sec:glm}. SNNs are equipped with $N = N_H + N_V$ neurons, where the number $N_V$ of output neurons is equal to the number of classes. For GLM-based SNNs, we assume a fully connected topology; while for SRM-based SNNs, we follow \cite{kaiser2020decolle} and consider a layered architecture. In particular, for SRM, we adopt either a two-layer fully connected architecture with each layer having $N_H/2$ neurons, or, for reference,  the convolutional architecture detailed in \cite{kaiser2020decolle}. Note that the latter generally requires non-local updates to deal with shared parameters in the convolutional kernels.


In Table \ref{tab:comparison-mnist}, we compare the test accuracies obtained on MNIST-DVS with both models after the sequential presentation of $100,000$ training examples, each selected (with replacement) from the training set. With DECOLLE, examples are presented in \textit{mini-batches} of size $100$, while the GLM-based SNN processes one at a time in an online fashion. As seen in Table~\ref{tab:comparison-mnist}, the SRM model with a convolutional architecture outperforms all other solutions, as long as one can accomodate the necessary number of hidden neurons and operate as a sufficiently high sampling rate. In contrast, in resource-constrained environments, when the sampling period and number of hidden neurons decreases, the performance of SRM-based training is significantly affected, while GLM-based solutions prove more robust. This suggests that the use of a global learning signal along with random sampling in GLM-based SNNs is less affected than local feedback signals by a reduction in model size and sampling rate. 



Next, we compare the training convergence speed of both models. To this end, we train the SNNs on the DVSGesture dataset. We set $N_H = 256$ and consider the same systems described above, to which we add an SNN trained with DECOLLE by presenting training examples one by one, i.e., with mini-batches of size one. We report the evolution of the training accuracies after training over $25,000$ examples. We note that training an SRM with DECOLLE is generally faster than training a GLM model as long as a sufficiently large mini-batch size is used. This suggests that, in this case, local per-neuron error signals can provide more informative updates than global error signals driven by random sampling.

\section{Conclusions} \label{sec:conclusion}
In this paper, we have reviewed the state of the art on the use of SNNs as processors  of spatio-temporal information in spiking signals. To that end, we have described SNNs as RNNs, and derived algorithms to train SRM and GLM-based SNN models. Based on neuromorphic datasets, we have provided insights on the learning performance and convergence under both models.





\bibliographystyle{IEEEtran}
\bibliography{ref}


\end{document}